# Adaptive ToR: Complexity-Aware Tree-Based Retrieval for Pareto-Optimal Multi-Intent NLU

[Hee-Kyong Yoo, Wonbae Kim, Hyocheol Ahn]

*[Data Science Lab Co., Ltd., Seoul, Korea]*

[hkyoo@dataslab.co.kr, wonkim@dataslab.co.kr, hcahn@dataslab.co.kr]

## Abstract

*Multi-intent natural language understanding requires retrieval systems that simultaneously achieve high accuracy and computational efficiency, yet existing approaches apply either uniform single-step retrieval that compromises recall or fixed-depth hierarchical decomposition that introduces excessive latency regardless of query complexity. This paper proposes Adaptive Tree-of-Retrieval (Adaptive ToR), a complexity-aware retrieval architecture that dynamically configures retrieval topology based on linguistic and structural query characteristics. The system integrates four components: (1) a Query Tree Classifier that computes a Query Complexity Index from weighted linguistic signals to route each query to either a rapid single-step path or an adaptive-depth hierarchical path; (2) a Tree-Based Retrieval module that recursively decomposes complex queries into focused sub-queries calibrated to predicted complexity; (3) an Adaptive Pruning Module employing two-stage filtering combining quantitative similarity gating with semantic relevance evaluation to suppress exponential node growth; and (4) a Retrieval Reranking Layer featuring a deduplicator-first pipeline and global LLM rescoring to ensure production efficiency. Evaluation on the NLU++ benchmark (2,693 multi-intent queries across Banking and Hotel domains) yields 29.07% Subset Accuracy and 71.79% Micro-F1, representing a 9.7% relative improvement over fixed-depth baselines, it simultaneously reduces latency by 37.6%, LLM invocations by 43.0%, and token consumption by 9.8%. Depth-wise analysis reveals that 26.92% of queries are resolved within three seconds (2.45 s mean latency) via single-step routing (d=0: 37.9% Subset Accuracy, 74.8% Micro-F1), while token consumption scales by a factor of 4.9 across depths, validating complexity-aware resource allocation and establishing Pareto-optimal balance across accuracy, latency, and computational efficiency within a unified architecture.*

## 1. Introduction

Natural Language Understanding (NLU) constitutes a foundational component of human-centric AI, enabling systems to interpret complex user intents and deliver contextually precise responses. As conversational interfaces proliferate across virtual assistants and enterprise applications, the demand for NLU systems that achieve both high accuracy and scalability has intensified—particularly in big data environments where heterogeneous queries must be processed under stringent latency constraints. Large Language Models (LLMs) substantially advance these capabilities, yet modern workloads increasingly involve multi-intent queries requiring simultaneous interpretation of multiple user goals within a single utterance, a challenge that exposes the limitations of traditional single-step retrieval paradigms [1].

Retrieval-Augmented Generation (RAG) addresses a fundamental limitation of purely parametric models by dynamically incorporating external evidence during inference, thereby reducing hallucinations and improving factual grounding [2]. However, standard RAG employs uniform retrieval strategies that compromise recall on compositional or multi-hop queries requiring hierarchical reasoning [3]. Tree-based extensions such as Tree-of-Retrieval RAG (ToR-RAG) [3] mitigate this limitation by decomposing complex queries into hierarchical sub-queries, expanding evidence coverage across multiple reasoning dimensions. Nevertheless, these approaches apply fixed-depth expansion regardless of query complexity, leading to systematic over-processing and introducing excessive computational overhead on simple queries that derive no benefit from hierarchical decomposition.

The practical consequences are significant: in production NLU systems, simple requests (e.g., "What is my account balance?") coexist with complex, multi-faceted instructions (e.g., "Compare interest rates between checking and savings accounts, then recommend the better option for emergency funds"). Applying uniform retrieval depth to both query types inefficiently allocates resources to trivial requests while potentially failing to adequately address genuinely complex ones. Recent research on adaptive systems [4] has emphasized query-aware resource allocation as essential for Pareto optimality, yet existing methods [4,5] primarily optimize post-retrieval reranking rather than the structural topology of retrieval itself. This gap motivates the development of a complexity-aware gating mechanism that predicts optimal retrieval depth prior to execution, thereby enabling proactive structural adaptation.

We propose Adaptive ToR, a complexity-aware retrieval architecture that dynamically configures retrieval topology based on query characteristics. The system integrates four components: (1) a Query Tree Classifier that routes queries to appropriate depth paths; (2) a Tree-based Retrieval module with recursive decomposition calibrated to predicted complexity; (3) an Adaptive Pruning Module employing two-stage filtering to suppress exponential node growth; and (4) a Retrieval Reranking Layer to ensure production-grade efficiency. Experimental evaluation on the NLU++ benchmark [6] demonstrates 29.07% Subset Accuracy and 71.79% Micro-F1, a 9.7% relative improvement over fixed-depth baselines, while reducing latency by 37.6% and LLM invocations by 43.0%.

This work makes five principal contributions to complexity-aware retrieval for multi-intent NLU: (1) a production-oriented Retrieval Reranking Layer that reduces latency by 37.6% and token consumption by 9.8% compared to fixed-depth approaches; (2) depth-wise performance validation empirically confirming correct query stratification by the Query Tree Classifier; (3) token-based cost quantification converting efficiency gains into operational savings; (4) a tri-objective Pareto optimality framework extending evaluation to accuracy, latency, and cost dimensions; and (5) contextualization within the broader adaptive retrieval research landscape, demonstrating that complexity-aware retrieval instantiates the design principle of adapting system behavior to user needs.

The Adaptive ToR architecture was initially presented in the first author's doctoral dissertation [7], which reported aggregate performance metrics. The present manuscript advances beyond that preliminary work by introducing the redesigned Retrieval Reranking Layer (contribution 1), providing the depth-wise mechanism validation and cost analysis absent from the dissertation (contributions 2–3), and establishing the Pareto optimality framework and research positioning (contributions 4–5). Concurrent work by the authors [8] has explored a complementary direction—lightweight, LLM-free query decomposition via embedding-space geometric segmentation—demonstrating that effective decomposition can also be achieved without generative inference for latency-sensitive cloud deployments.

The remainder of this paper is organized as follows. Section 2 reviews related work in RAG systems, tree-based retrieval, adaptive mechanisms, and multi-intent NLU. Section 3 details the Adaptive ToR architecture. Section 4 describes the experimental setup, including the dataset, evaluation metrics, implementation details, and baseline systems. Section 5 presents experimental results, depth-wise analysis, cost quantification, and Pareto frontier evaluation. Section 6 concludes with implications and future directions.

## 2. Related Work

### 2.1 Evolution of Retrieval-Augmented Generation Systems

Retrieval-Augmented Generation (RAG) has fundamentally transformed knowledge-intensive natural language processing by enabling Large Language Models (LLMs) to incorporate external evidence during inference. The foundational RAG architecture introduced by Lewis et al. [2] established the canonical retrieval pipeline: dense retrieval of top-k documents via semantic similarity, followed by concatenation with the input query for LLM generation. This approach demonstrated substantial improvements in open-domain question answering and fact verification tasks by grounding model outputs in verifiable external sources. However, as RAG systems have been deployed in increasingly complex real-world scenarios, particularly in interactive applications requiring real-time responsiveness, limitations of the standard single-step retrieval paradigm have become apparent.

A recent survey have documented a paradigm shift toward adaptive and agentic RAG architectures that dynamically modulate retrieval behavior based on query characteristics. Singh et al. [1] present a comprehensive overview of agentic RAG systems, highlighting the transition from static retrieval pipelines to autonomous agents capable of iterative reasoning and self-correction. Recent advances have further expanded RAG capabilities: Izacard et al. [9] introduced Atlas, demonstrating that few-shot learning with retrieval augmentation significantly improves generalization across diverse tasks. This design philosophy aligns with the growing role of LLM-based automation in adaptive information processing, where models serve dual roles as both generators and evaluators, a capability that our Adaptive Pruning Module similarly exploits through its LLM-as-a-Judge strategy [10].

Jeong et al. [4] proposed Adaptive-RAG, which classifies query complexity to select among three retrieval strategies: no retrieval for simple factual queries, single-step retrieval for moderate complexity, and iterative multi-step retrieval for complex reasoning tasks. Similarly, Corrective RAG (CRAG) [5] introduced a self-reflective mechanism that evaluates retrieved document relevance and triggers corrective actions, including query refinement and knowledge base expansion, when initial retrieval quality is insufficient. These studies demonstrate that query-aware adaptation is essential for achieving Pareto optimality in production-grade NLU systems, yet they primarily operate at the post-retrieval reranking level rather than reconfiguring the structural topology of the search space itself.

More recently, the integration of RAG with agentic frameworks has enabled autonomous reasoning trajectories. Self-RAG [11] equips LLMs with reflection tokens that dynamically decide when to retrieve, which passages to use, and how to self-critique generated outputs, effectively converting the model into an autonomous agent capable of iterative self-improvement. Graph-RAG [12] extends this paradigm by constructing knowledge graphs from retrieved documents and performing multi-hop reasoning over graph structures, demonstrating improved performance on complex analytical queries. However, these graph-based methods often rely on static knowledge structures that cannot adapt to the varying complexity of individual queries. Our work addresses this limitation by introducing complexity-aware tree-based retrieval, which dynamically adjusts both the depth and breadth of hierarchical decomposition based on a quantitative assessment of query difficulty performed before retrieval begins.

## 2.2 Tree-Based Retrieval and Hierarchical Decomposition

The emergence of tree-based retrieval architectures represents a significant advance in handling compositional and multi-hop queries. Our prior work on Tree-of-Retrieval RAG (ToR-RAG) [3] demonstrated that hierarchical query decomposition substantially improves evidence coverage for multi-intent NLU tasks by recursively breaking down complex queries into semantically focused sub-queries. While query expansion techniques such as Query2doc [13] enhance retrieval by enriching sparse queries with LLM-generated pseudo-documents, they operate at the lexical level without restructuring the retrieval topology. In contrast, tree-based approaches align with recent findings showing that interleaving retrieval with chain-of-thought reasoning significantly improves performance on knowledge-intensive multi-step questions [14], and that explicit decomposition strategies effectively narrow the compositionality gap in language models [15]. Each node in the resulting tree structure targets a distinct information need, enabling comprehensive retrieval across multiple reasoning dimensions. Experimental results on the MultiHop-RAG benchmark showed that ToR-RAG achieved improvements of +6.42 in Exact Match and +6.04 in F1 compared to baseline approaches, with optimal performance at depth-3, validating the effectiveness of tree-based decomposition for complex question processing.

Concurrent work has explored similar hierarchical paradigms, focusing on entity-level indexing, topic-based organization, and domain-specific applications. Li et al. [16] introduced CFT-RAG, which leverages entity hierarchies and Cuckoo Filters to accelerate physical search operations while reducing hallucinations through structured knowledge organization. Fatehkia et al. [17] proposed T-RAG, employing a tree-structured knowledge base organized by topic hierarchies that enables more efficient traversal of domain-specific information spaces. Yang and Huang [18] applied tree-based RAG structures to medical test recommendation in HiRMed, demonstrating domain-specific applicability of hierarchical retrieval. While these methods demonstrate the value of hierarchical organization, they largely assume fixed tree structures defined either by domain ontologies or predetermined decomposition depths. This assumption breaks down in big data analytics environments where query complexity varies dramatically, from simple account balance inquiries to complex multi-faceted comparative analyses, necessitating dynamic depth adaptation rather than uniform expansion policies.

Recent work in hardware-aware retrieval systems has further highlighted the importance of computational efficiency in hierarchical retrieval. Chen et al. [19] proposed REIS, an in-storage processing architecture for retrieval systems that minimizes data movement overhead and achieves 2.3 times energy efficiency improvements compared to conventional designs. Their findings underscore that retrieval depth directly impacts both latency and energy consumption, making adaptive depth control not merely an optimization but a fundamental requirement for sustainable deployment of RAG systems at scale. Our Adaptive ToR framework operationalizes these insights by introducing the Query Tree Classifier (QTC), which predicts optimal retrieval depth a priori based on linguistic complexity signals, thereby eliminating unnecessary tree expansion for simple queries while preserving deep reasoning capabilities for genuinely complex tasks. Our concurrent work on ToR-Lite [8] offers a complementary LLM-free alternative, discussed further in Section 2.5.

## 2.3 Adaptive Systems and Information Retrieval

The concept of adaptive information retrieval has deep roots in human-computer interaction and information-seeking research, which emphasize tailoring system behavior to user intent and context. Recent work has demonstrated the value of complexity-aware adaptation across diverse application domains, establishing a critical design principle directly relevant to our research: adaptive depth control based on observable signals significantly improves system performance by eliminating unnecessary overhead for simple requests while ensuring comprehensive coverage for complex inquiries. Our Query Complexity Index (QCI), derived from linguistic signals such as WH-terms, conjunctions, and comparative expressions, operationalizes this principle by enabling proactive resource allocation before expensive retrieval operations commence.

Beyond these observations, broader research on adaptive query processing has explored dynamic optimization strategies. Khattab et al. [20] introduced the Demonstrate-Search-Predict (DSP) framework, which composes retrieval and language models for knowledge-intensive NLP by adaptively structuring the retrieval-reasoning pipeline based on task requirements, demonstrating that modular composition of retrieval and generation stages can achieve significant performance improvements. Oruche et al. [21] surveyed recent advancements in human-centered dialog systems, noting that modern conversational agents must support mixed-initiative interaction where both user complexity and system capability are dynamically balanced. These cross-disciplinary findings collectively validate our core thesis: pre-execution complexity assessment enables structural adaptation that yields superior Pareto trade-offs compared to post-hoc reranking approaches. Our Adaptive ToR framework embodies these principles by implementing a dual-path architecture—simple queries bypass hierarchical decomposition entirely via the Simple Path (d=0), while complex queries trigger adaptive tree expansion via the Tree Path (d $\in$ {1,2,3}) calibrated to the predicted complexity level, ensuring that computational resources are allocated proportionally to user intent complexity.

## 2.4 Multi-Intent Natural Language Understanding

Multi-intent NLU represents a significant departure from traditional single-intent classification, requiring systems to simultaneously recognize and interpret multiple user goals within a single utterance. Early approaches treated this as a multi-label classification problem, employing joint slot-filling and intent detection models based on recurrent neural networks. However, these methods struggle with compositional queries that contain nested logical structures or require evidence synthesis across multiple information sources.

Recent work has shifted toward hierarchical and decomposition-based strategies for multi-intent understanding. These approaches explicitly model the compositional structure of complex queries where one intent constrains or qualifies another (e.g., "Compare X and Y, then recommend the better option for Z"), demonstrating that tailoring the retrieval space to the specific semantic structure of the query significantly improves multi-intent classification accuracy by reducing noise from irrelevant evidence.

The NLU++ benchmark [6], which we employ for evaluation, exemplifies the challenges of modern multi-intent understanding. With an average of 2.01 intents per query and 62 intent classes spanning Banking and Hotel domains,

NLU++ requires systems to handle diverse compositional structures: conjunctive intents ("A and B"), comparative intents ("A vs. B, which is better?"), and sequential intents ("First A, then B"). Standard RAG achieves only 27.29% Subset Accuracy on this benchmark, while fixed-depth ToR-RAG attains 26.51% [3], underscoring the necessity of adaptive depth control to match retrieval strategy with query structure.

## 2.5 Research Gaps and Positioning of Adaptive ToR

Despite the substantial advances reviewed above, existing approaches exhibit three critical limitations that Adaptive ToR is specifically designed to address.

The first gap pertains to post-hoc optimization versus structural adaptation. Prior adaptive RAG systems such as Adaptive-RAG [4] and CRAG [5] perform complexity-aware routing after retrieval candidates are generated, optimizing reranking strategies but not the retrieval topology itself. In contrast, our Query Tree Classifier (QTC) performs proactive structural adaptation by adjusting tree depth before retrieval begins, fundamentally reconfiguring the search space rather than merely reordering its outputs.

The second gap involves fixed-depth hierarchical expansion. Tree-based methods including ToR-RAG [3] and CFT-RAG [16] demonstrate the value of hierarchical decomposition but apply uniform expansion depths regardless of query complexity, leading to systematic over-processing for simple queries and potential under-processing for genuinely complex ones. Our complexity-aware approach addresses this limitation by dynamically calibrating depth ($d \in \{0,1,2,3\}$) based on the Query Complexity Index, adapting the retrieval topology to the intrinsic difficulty of each individual query. The effectiveness of this approach is empirically validated in Section 5.

The third gap relates to the lack of unified Pareto evaluation. Existing work typically reports accuracy or efficiency metrics in isolation, lacking a rigorous framework for joint optimization. We establish a Pareto optimality methodology that simultaneously quantifies accuracy (Subset Accuracy, Micro-F1) and efficiency (latency, token cost, LLM invocations) on a unified frontier, enabling systematic evaluation of whether complexity-aware adaptation achieves gains on both dimensions rather than trading one for the other.

Adaptive ToR is the second work in the Tree-of-Retrieval research lineage: ToR-RAG [3] established the effectiveness of hierarchical decomposition with fixed-depth expansion; Adaptive ToR introduces complexity-aware dynamic depth control; and the concurrent ToR-Lite [8] completes this progression by demonstrating that effective decomposition can be achieved through lightweight, LLM-free geometric methods, occupying a Pareto-dominant position for latency-sensitive cloud deployments. While Adaptive ToR and ToR-Lite share the same research lineage, they address structurally orthogonal problems—Adaptive ToR targets complexity-aware intent classification evaluated on NLU++, whereas ToR-Lite targets LLM-free multi-hop retrieval decomposition evaluated on MultiHop-RAG [8]—and are not interchangeable.

By synthesizing insights from adaptive RAG, tree-based retrieval, and multi-intent NLU, Adaptive ToR introduces a complexity-aware retrieval approach. The architecture dynamically self-configures to match the intrinsic complexity of each query, targeting Pareto-optimal performance across heterogeneous workloads where simple requests coexist with complex multi-intent instructions.

## 3. Proposed Architecture: Adaptive ToR

This section presents Adaptive Tree-of-Retrieval (Adaptive ToR), a Pareto-optimal retrieval architecture designed to achieve high intent coverage while maintaining computational efficiency for multi-intent query processing. Standard single-step retrieval applies flat retrieval regardless of query complexity, whereas fixed-depth tree-based approaches uniformly expand all queries to maximum depth, introducing excessive latency for simple queries. Adaptive ToR addresses this gap by dynamically configuring the retrieval topology based on query characteristics—routing simple queries through a rapid single-step path while reserving hierarchical decomposition for genuinely complex multi-intent queries.

The architecture comprises four core components, illustrated in Figure 1: (1) a Query Tree Classifier (QTC) that computes a Query Complexity Index (QCI) to route queries to appropriate depth paths; (2) a Tree-Based Retrieval Module with recursive decomposition calibrated to predicted complexity; (3) an Adaptive Pruning Module (APM) employing two-stage filtering to suppress exponential node growth; and (4) a Retrieval Reranking Layer (RRL) ensuring production-grade efficiency through deduplication and global rescoring. Each component is detailed in the following subsections.

## 3.1 System Overview

Adaptive ToR implements a Pareto-optimal dual-path architecture designed to efficiently handle the significant variance in query difficulty within real-world NLU workloads. Unlike conventional approaches that process all requests uniformly, the system eliminates structural inefficiencies by dynamically allocating execution paths based on query-specific complexity.

As illustrated in Figure 1, the execution flow begins with the Query Tree Classifier (QTC), which evaluates the structural intricacy of the input natural language query (NLQ) using weighted linguistic markers. The QTC routes the query into one of two specialized pipelines:

**Simple Path (d = 0):** Routine, single-intent queries bypass both hierarchical expansion and the Retrieval Reranking Layer. They undergo single-step dense retrieval directly against the knowledge base (top-k) and deliver evidence directly to the downstream application, minimizing latency for straightforward queries.

**Tree Path (d ∈ {1, 2, 3}):** Complex, multi-intent queries are routed to the Tree-of-Retrieval (ToR) for hierarchical sub-query decomposition. A key component of the QTC is the Tree-Depth Controller, which selects the maximum depth d before decomposition begins. This preemptive resource allocation ensures that computational power is scaled only when structural complexity warrants it, preventing the over-processing common in fixed-depth models.

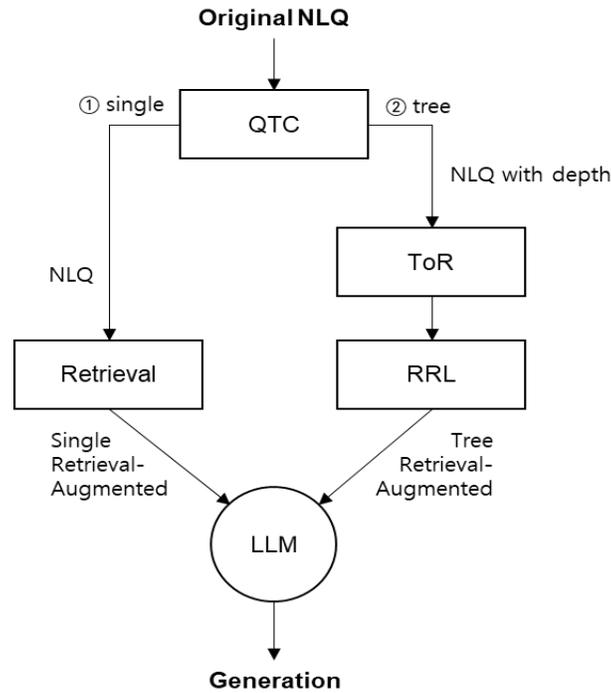

**Figure 1. Overall Architecture of Adaptive ToR System (presented in [8, Figure 2])**

**End-to-End Workflow**

(1) Analysis & Routing: Upon receiving a user query (NLQ), the QTC computes a Query Complexity Index (QCI). If QCI < 0.10 and no conjunctions or comparisons are detected, the query follows the Simple Path. Otherwise, it is assigned a predicted depth d and routed to the Tree Path.

(2) Retrieval & Pruning: The Simple Path executes a single-step retrieval and bypasses further processing. The Tree Path recursively decomposes the query into sub-queries, during which the Adaptive Pruning Module (APM) filters low-relevance branches via two-stage gating to mitigate computational overhead.

(3) Consolidation: For Tree Path queries, the Retrieval Reranking Layer (RRL) aggregates evidence from all tree nodes, performing deduplication and global rescoring using a Reranker LLM. This ensures that only high-relevance, non-redundant information is passed to the downstream application.

This dual-path design achieves Pareto dominance by providing rapid responses for routine tasks while enabling deep, structured retrieval for complex reasoning, effectively balancing precision and computational cost.

## 3.2 Query Tree Classifier (QTC)

The QTC quantifies the linguistic complexity of an input query to determine the optimal execution path and retrieval depth. It operates through two integrated modules: the Query Complexity Router, which performs binary path classification (Simple vs. Tree Path), and the Tree Depth Controller, which selects adaptive depth for Tree Path queries, as illustrated in Figure 2.

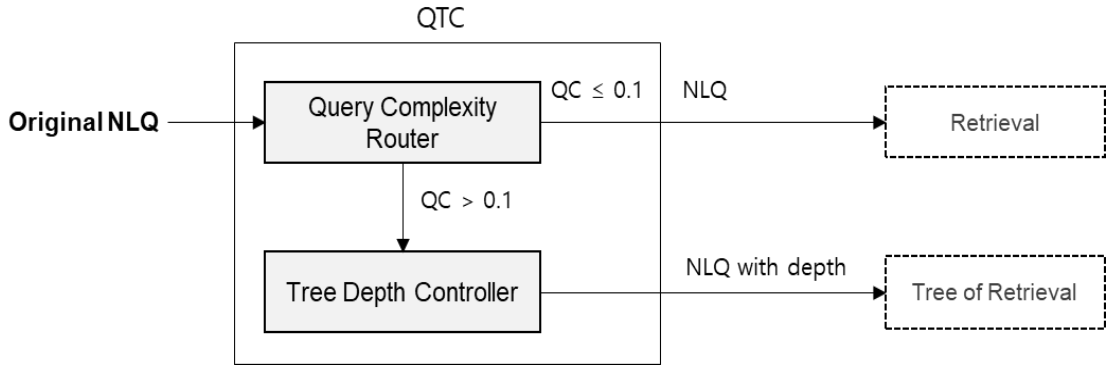

**Figure 2. Query Tree Classifier (QTC)**

### 3.2.1 Query Complexity Router

The router's primary objective is to proactively identify unambiguous single-intent queries using surface-level linguistic cues, thereby enabling them to bypass the computationally intensive Tree Path. To minimize false negatives (i.e., misclassifying complex queries as simple), the router maintains a conservative gating policy: any query exhibiting structural ambiguity or multi-faceted indicators is routed to the Tree Path.

The router computes a Query Complexity Index (QCI) by extracting and weighting five linguistic features, as summarized in Table 1.

The QCI is computed as the weighted sum of five normalized signal values:

$$QCI(Q) = \Sigma_i \, w_i \cdot s_i(Q)$$

where each signal value $s_i(Q) \in [0, 1]$. The WH-term, Conjunction, Comparison, and Sequence signals are binary indicators (1 if the corresponding linguistic pattern is detected, 0 otherwise). The Length signal is computed as min(token_count / 25, 1), where 25 tokens serves as the normalization threshold. Queries are routed to the Simple Path if QCI < τ_simple and no conjunctions or comparisons are detected, where τ_simple = 0.10. All remaining queries are assigned to the Tree Path.

**Table 1.** QCI Signal Definitions and Weights

| Signal | Weight | Theoretical Basis |
|---|---|---|
| WH-term | 0.25 | Presence of interrogative words |
| Conjunction | 0.20 | Coordinating/subordinating conjunction (and, or, but, while) |
| Comparison | 0.20 | Comparative/Contrastive terms (compare, versus, better, difference) |
| Sequence | 0.15 | Temporal/sequential indicators (first, then, after, before, next) |
| Length | 0.20 | Query token count normalized by threshold (25 tokens) |

### 3.2.2 Tree Depth Controller (TDC)

For queries assigned to the Tree Path, the TDC selects the maximum exploration depth ($d \in \{1, 2, 3\}$), which functions as a critical parameter for regulating the computational budget. The depth selection is based on structural complexity indicators, including conjunction count, comparison markers, and the Query Complexity Index (QCI) defined in Section 3.2.1. To prevent an exponential explosion of the search space, the TDC strictly caps the exploration depth at three levels, ensuring a manageable maximum of eight leaf nodes in worst-case scenarios (i.e., when no pruning occurs). This constraint is informed by prior work indicating that extending depth beyond three levels typically results in significant performance degradation [7].

For Tree Path queries, the TDC determines the maximum depth d according to the following rules. Queries initially classified as Tree mode undergo an additional LLM-based complexity assessment that assigns a semantic difficulty level $L \in \{Low, Mid, High\}$ based on the original query, retrieved context snippets, and the initial mode classification. The depth is then assigned as:

d = 1 if L = Low, d = 2 if L = Mid, d = 3 if L = High.

Queries classified as Simple or Hybrid mode by the initial router are assigned d = 0 and bypass tree expansion entirely, following the Simple Path. This mapping is summarized in Table 2.

**Table 2:** Depth Assignment Rules

| Initial Mode | Semantic Level | Depth (d) | Path |
|---|---|---|---|
| Simple | - | 0 | Simple Path |
| Hybrid | - | 0 | Simple Path |
| Tree | Low | 1 | Tree Path |
| Tree | Mid | 2 | Tree Path |
| Tree | High | 3 | Tree Path |

The LLM-based complexity assessment is invoked only for queries classified as Tree mode by the rule-based router, minimizing additional computational overhead. This selective invocation ensures that the cost of complexity estimation (C_QCC) remains bounded, as approximately 73% of queries in the NLU++ dataset are routed to the Tree Path and require this additional assessment, while the remaining 27% bypass it entirely via the Simple Path.

## 3.3 Tree-of-Retrieval (ToR)

The ToR module serves as the core component for processing complex, multi-intent queries by generating a hierarchical structure of decomposed semantic units. This process ensures comprehensive coverage of compositional requirements by recursively expanding the original query into sub-queries until the target depth (d) is reached. As illustrated in Figure 3, the system facilitates a structured expansion where the root query (NLQ) is systematically branched into specialized decomposed queries (DQs). At each expansion stage, the APM (Adaptive Pruning Module) is integrated to perform node-wise candidate filtering, ensuring that only high-relevance nodes propagate to the next level. This integration of expansion and pruning prevents the search space from growing unmanageably while maintaining deep reasoning capabilities.

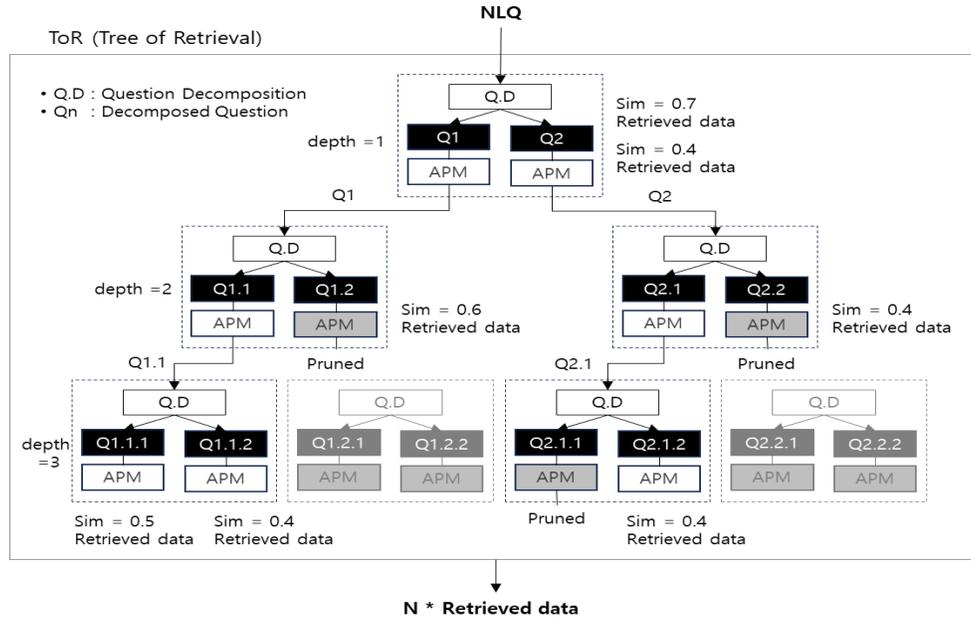

**Figure 3. Tree-of-Retrieval (ToR) expansion with node-wise pruning (presented in [8, Figure 1])**

### 3.3.1 Question Decomposition (Q.D)

Question Decomposition is the fundamental operation that subdivides a high-level query into its minimal processable semantic units. The system recursively generates two (2) sub-queries at each level, adhering to strict quality constraints: each sub-query must be specific (clear interrogative forms), relevant (alignment with the parent query), non-redundant (minimal semantic overlap), complete (covering all parent intents), and appropriately granular (isolating core information axes).

### 3.3.2 Adaptive Pruning Module (APM)

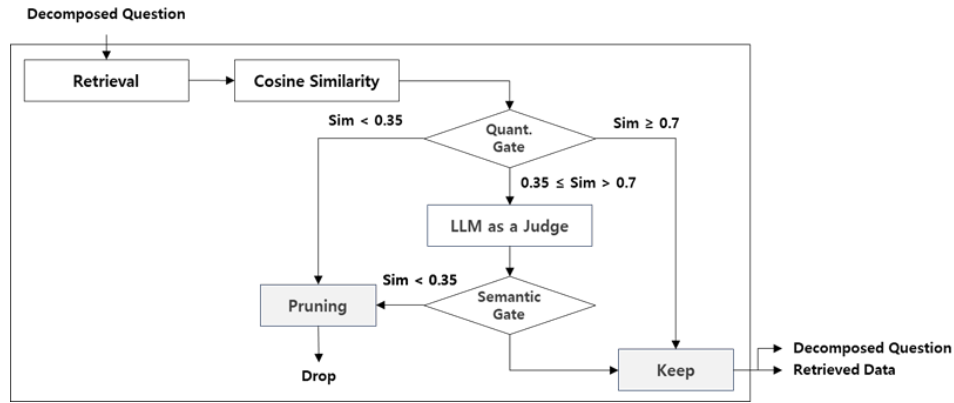

**Figure 4. Two-stage Adaptive Pruning Module (APM)**

To

To mitigate the exponential surge in computational costs during tree expansion, the APM functions as a node-local filtering mechanism through a precision-driven two-stage pipeline.

As illustrated in Figure 4, the process begins with Stage 1 (Quantitative Gate), where candidates are classified using low-cost statistical scores such as cosine similarity. High-relevance candidates (sim ≥ 0.70) are directly retained, while low-relevance candidates (sim < 0.35) are immediately discarded. Borderline cases (0.35 ≤ sim < 0.70) are passed to Stage 2 (Semantic Gate) for an in-depth "LLM-as-a-Judge" evaluation [10]. This qualitative assessment resolves complex contextual nuances that numerical scores alone cannot distinguish. By optimizing the candidate pool at the node level, the APM effectively suppresses the escalation of retrieval and adjudication costs in subsequent phases.

## 3.4 Retrieval Reranking Layer (RRL)

The RRL consolidates fragmented evidence collected through the Tree Path. Unlike the Simple Path, the Tree Path aggregates candidates from multiple sub-query nodes, which often results in data redundancy and a discrepancy between local sub-query relevance and global alignment with the original query. The preliminary Adaptive ToR architecture [7] addressed this consolidation through an Intent Aggregator LLM (IAL) that first performed Cross-Encoder-based reranking on the full candidate pool, then applied Maximal Marginal Relevance (MMR) filtering to suppress redundancy, and finally mapped surviving candidates to an intent catalog via embedding similarity thresholding. While effective, this rerank-then-deduplicate ordering introduced two inefficiencies: (1) the Cross-Encoder processed redundant passages that would subsequently be removed, increasing token consumption; and (2) reranking fragmented sub-query evidence without first consolidating it led to context fragmentation during relevance evaluation. The RRL addresses both limitations by inverting the pipeline order: deduplication precedes reranking, and global rescoring replaces sub-query-level evaluation.

### 3.4.1 Deduplicator and Global Reranking

The RRL introduces a deduplicator-first pipeline, a key methodological departure from prior work [7]. The system first employs a Deduplicator to merge the candidate pool, removing redundant passages and enhancing information density. This preemptive deduplication reduces computational costs by minimizing the number of passages requiring expensive Reranker LLM processing and prevents over-representation of specific evidence that could bias downstream intent classification.

Following deduplication, the Reranker LLM (LLaMA-3.1-8B-Instruct) rescores each remaining candidate passage against the original natural-language query (not decomposed sub-queries), yielding a global relevance score. This global rescoring strategy achieves two objectives: (1) it mitigates hallucinations by grounding the reranking process in the user's overarching intent rather than sub-query fragments, and (2) it maintains full query context during relevance evaluation, ensuring that evidence from all tree branches receives equal consideration and preventing important information from being overlooked in lengthy candidate pools.

Efficiency gains: The deduplicator-first approach is designed to reduce token consumption and latency by eliminating redundant passages before Reranker LLM processing. Empirical validation of these efficiency gains is presented in Section 5.2.

### 3.4.2 Top-K Selection with Dual Thresholding

Finally, the Top-K Selection module applies a dual-threshold global cutoff to the rescored candidates, passages are retained if they rank within the top-10 or exceed a relevance score of 0.70; the final candidate set is the union of both criteria, capped at 10 passages. This design ensures that high-quality evidence is never excluded due to arbitrary rank cutoffs while simultaneously capping context size to prevent token budget overflow. The dual-threshold mechanism provides predictable latency and efficient token utilization for downstream NLU applications, with the top-10 limit guaranteeing a maximum context window regardless of query complexity.

## 4. Experimental Setup

## 4.1 Dataset

We evaluated all systems on the NLU++ benchmark, a curated dataset designed for multi-intent query understanding in conversational AI domains. The dataset comprises 2,693 intent-labeled natural language queries spanning two domains: Banking (account management, transactions, card services) and Hotel (reservations, amenities, customer service). From an initial pool of 3,080 queries, 362 unlabeled instances and 25 duplicates were removed to yield the final evaluation set. Each query is annotated with one or more intent labels from a taxonomy of 62 distinct intent classes. The average number of intents per query is 2.01, with a distribution ranging from single-intent queries (e.g., "What is my account balance?") to complex multi-intent requests (e.g., "Compare the interest rates for savings accounts, then recommend the best option and show me how to apply online"). Note that NLU++ evaluates multi-intent classification in task-oriented dialogue, which is structurally distinct from the multi-hop retrieval decomposition evaluated using MultiHop-RAG in our concurrent work [8].

## 4.2 Evaluation Metrics

We adopt three standard metrics for multi-label classification. Subset Accuracy (primary metric) is the proportion of queries for which the system correctly predicts the exact set of intents. This strict metric penalizes partial matches and reflects real-world NLU requirements where missing even one intent can degrade user experience. Micro-F1 is the F1 score computed globally across all intent predictions, emphasizing performance on frequent classes. Macro-F1 is the arithmetic mean of per-class F1 scores, weighting each of the 62 intent classes equally regardless of frequency. This metric captures performance on rare intents that Micro-F1 may underrepresent, providing a complementary view of classification quality across the full intent taxonomy. In addition to accuracy, we measure query-processing latency (wall-clock time from query input to evidence retrieval completion, excluding final answer generation) and LLM API call count (total number of language model invocations per query, reflecting operational cost).

## 4.3 Implementation Details

### 4.3.1 Language Model

All LLM-based components use LLaMA-3.1-8B-Instruct served via the Ollama framework with FP16 inference precision. Query decomposition operates at temperature = 0.3 to allow moderate lexical diversity in sub-query generation, while the adaptive pruning judge uses temperature = 0.1 to encourage deterministic relevance assessments. The same model is also used for final intent classification and downstream answer generation; the latter falls outside the scope of the present evaluation.

### 4.3.2 Vector Database

Weaviate v1.25 with HNSW indexing configured at M = 32 (maximum neighbor connections per node) and efConstruction = 200 (dynamic candidate list size during index building). These parameters balance retrieval recall against index construction cost. Document embeddings were generated using all-mpnet-base-v2 (768-dimensional dense vectors).

### 4.3.3 Retrieval Configuration

Both paths retrieve the top-32 (k = 32) most similar passages per retrieval operation. The Simple Path performs a single retrieval against the full knowledge base, yielding up to 32 candidate passages. The Tree Path retrieves top-32 passages per surviving node at each tree level; the Adaptive Pruning Module subsequently filters this candidate pool before propagation to the next depth level.

### 4.3.4 Reranking

Reranker LLM rescores all retrieved passages against the original query. Final candidate set consists of the top-10 passages or those with global relevance score $\geq 0.70$.

### 4.3.5 Hardware

NVIDIA RTX A6000 (48GB VRAM) GPU and 64-core AMD EPYC 7742 CPU. All timing measurements averaged over 10-fold cross-validation stratified by domain distribution to account for query complexity variance.

**4.4 Baseline Systems**

We compare Adaptive ToR against two representative retrieval architectures:

**Standard RAG**, a conventional single-step retrieval-augmented generation system. The input query is directly embedded and matched against the vector database to retrieve the top-32 most similar passages. This baseline represents the efficiency-first approach, prioritizing rapid response at the potential expense of completeness for complex queries.

**Fixed-depth ToR-RAG**, a hierarchical retrieval system with uniform tree expansion to depth d=3 for all queries. Every non-leaf node undergoes LLM-based decomposition into binary sub-queries, and retrieval proceeds independently at each leaf. This baseline represents the quality-first approach, applying maximum decomposition uniformly without query-adaptive optimization. It serves to isolate the benefit of complexity-aware depth control.

Both baselines use the same LLMs, embeddings, vector databases, and reranking modules as Adaptive ToR, ensuring that performance differences arise solely from architectural decisions rather than implementation artifacts.

As an additional reference point, we note that a BERT-base classifier (fine-tuned directly on NLU++ without any retrieval component) achieves only 15.04% Subset Accuracy on this benchmark, substantially below all RAG-based systems. This result underscores that the multi-intent, slot-rich structure of NLU++ exceeds the capacity of purely parametric classification, motivating the retrieval-augmented approaches evaluated in this study. Since BERT-base operates without retrieval and thus differs fundamentally in architecture, we do not include it as a primary baseline but report this figure to contextualize the difficulty of the benchmark.

# 5. Results

## 5.1 Overall performance comparison analysis

Table 2 summarizes the comparative performance of the three retrieval architectures on NLU++. Adaptive ToR demonstrates substantial improvements across all dimensions of the evaluation.

**Table 3:** Overall Performance Comparison on NLU++ Benchmark

| Model | Subset Acc (%) | Micro-F1 (%) | Macro-F1 (%) | Latency (s) | Ave Depth | LLM Calls |
|---|---|---|---|---|---|---|
| Adaptive ToR | 29.07 | 71.79 | 71.51 | 9.73 | 1.53 | 6.01 |
| ToR-RAG | 26.51 | 70.97 | 71.43 | 15.58 | 3.00 | 10.54 |
| Standard RAG | 27.29 | 63.16 | 67.71 | 5.61 | 0.00 | 2.0 |

**Key Findings:**

(1) Accuracy Improvements

Adaptive ToR achieves 29.07% Subset Accuracy, representing a +2.56 percentage-point (pp) absolute gain over Fixed-depth ToR (+9.7% relative improvement) and +1.78 pp over Standard RAG. The Micro-F1 improvement of +8.63 pp over Standard RAG (63.16% → 71.79%) confirms that adaptive decomposition substantially enhances intent coverage for compositional queries. Macro-F1 exhibits a consistent pattern: Adaptive ToR achieves 71.51%, compared to 71.43% for Fixed-depth ToR and 67.71% for Standard RAG. The +3.80 pp gain over Standard RAG indicates that the improvement is not confined to high-frequency intents but extends across the full 62-class taxonomy, including rare intent categories that Micro-F1 alone may underrepresent.

(2) Latency Reduction

Despite achieving higher accuracy than Fixed-depth ToR, Adaptive ToR reduces mean query-processing latency by 37.6% (15.58 s → 9.73 s). This counterintuitive result stems from the Query Tree Classifier's ability to bypass unnecessary hierarchical expansion for simple queries.

(3) Cost Efficiency

Adaptive ToR requires an average of 6.01 LLM calls per query, a 43.0% reduction compared to Fixed-depth ToR's 10.54 calls. The average tree depth of 1.53 (versus the fixed depth of 3.0) demonstrates effective complexity-aware resource allocation. Even when compared to Standard RAG's minimal 2.0 calls, Adaptive ToR requires approximately 3 times as many LLM calls; however, this additional cost is justified by the substantial accuracy gains and the practical necessity of handling complex multi-intent queries that single-step retrieval cannot adequately address.

These results confirm that complexity-aware adaptive retrieval reconciles the traditional accuracy-efficiency trade-off, simultaneously outperforming both baseline strategies in the joint objective space.

## 5.2 Depth-Wise Performance Validation

Table 3 presents depth-stratified performance metrics across d ∈ {0, 1, 2, 3}, providing empirical validation of the Query Tree Classifier's (QTC) complexity-aware routing mechanism—a granular analysis absent from the doctoral dissertation [7], which reported only aggregate metrics (29.07% Subset Accuracy, 9.73 s latency). This depth-wise breakdown constitutes a mechanism validation study that demonstrates how computational resources are allocated proportionally to query difficulty.

**Table 4:** Performance Metrics by Retrieval Depth (NLU++ Benchmark)

| Depth | Query (%) | Subset Acc. (%) | Micro-F1 (%) | Latency (s) | Ave. Prompt Tokens |
|---|---|---|---|---|---|
| d=0 | 26.92 (725) | 37.90 | 74.80 | 2.45 | 2,116 |
| d=1 | 1.56 (42) | 21.40 | 66.70 | 5.55 | 6,227 |
| d=2 | 62.46 (1,682) | 27.70 | 71.00 | 10.01 | 8,140 |
| d=3 | 9.06 (244) | 13.50 | 71.40 | 15.70 | 10,366 |
| Weighted Average | 100 (2,693) | 29.07 | 71.79 | 9.73 | 6,689 |

**Key Insights:**

(1) Complexity-cost alignment validated: Token consumption scales monotonically with assigned depth (2,116 → 10,366 tokens, a factor of 4.9), and latency increases correspondingly (2.45 s → 15.70 s). This monotonic progression confirms that the QTC correctly stratifies queries by intrinsic difficulty and that the Tree Depth Controller (TDC) allocates computational budgets proportionally to predicted complexity, validating the core hypothesis of complexity-aware retrieval.

(2) Fast-path effectiveness confirmed: The Simple Path (d=0) handles 26.92% of queries with the highest Subset Accuracy (37.90%), highest Micro-F1 (74.80%), and lowest latency (2.45 s)—achieving sub-3-second response times. This validates the dual-path architecture's efficiency gains: routine queries bypass hierarchical decomposition entirely, eliminating unnecessary computational overhead while maintaining superior accuracy.

(3) Counter-intuitive deep-decomposition pattern: At maximum depth (d=3), Subset Accuracy drops to 13.50% (lowest across all depths) while Micro-F1 remains high (71.40%, slightly exceeding d=2's 71.00%). This indicates that deep hierarchical decomposition successfully recovers partial intents for the most challenging queries despite failing to match all intents exactly (hence low Subset Accuracy). The 9.06% of queries routed to d=3 represent genuinely complex cases requiring maximum reasoning depth, justifying the computational investment.

(4) Token-based cost quantification: Token consumption is estimated using a character-count heuristic (prompt characters / 4), a widely adopted approximation sufficient for relative comparison between architectures. The weighted-average token consumption per query is 6,689 for Adaptive ToR, representing a 9.8% reduction compared to fixed-depth ToR-RAG's 7,417 tokens. The 4.9x token variation across depths (2,116 → 10,366) confirms that computational resources scale proportionally with query complexity, validating the efficiency of complexity-aware resource allocation.

**Mechanism validation.** These results empirically confirm that Adaptive ToR's complexity-aware routing prevents systematic over-processing: 26.92% of queries achieve optimal performance (d=0) without incurring the latency penalty of deeper expansion, while only 9.06% require maximum depth. The majority workload (62.46%) concentrates at d=2, indicating that the QTC successfully identifies a middle ground for moderately complex queries. This distribution demonstrates that the system dynamically self-configures retrieval topology to match query-specific demands, achieving a Pareto-optimal trade-offs and validating the architectural hypothesis absent from prior work [7].

## 5.3 Pareto Optimality Analysis

A core objective of Adaptive ToR is achieving Pareto optimality, a state where no alternative system can simultaneously outperform across all performance dimensions. Figure 5 presents a two-dimensional visualization comparing the three architectures on Micro-F1 versus average latency. The analysis extends beyond this two-dimensional projection to encompass LLM invocations and token consumption.

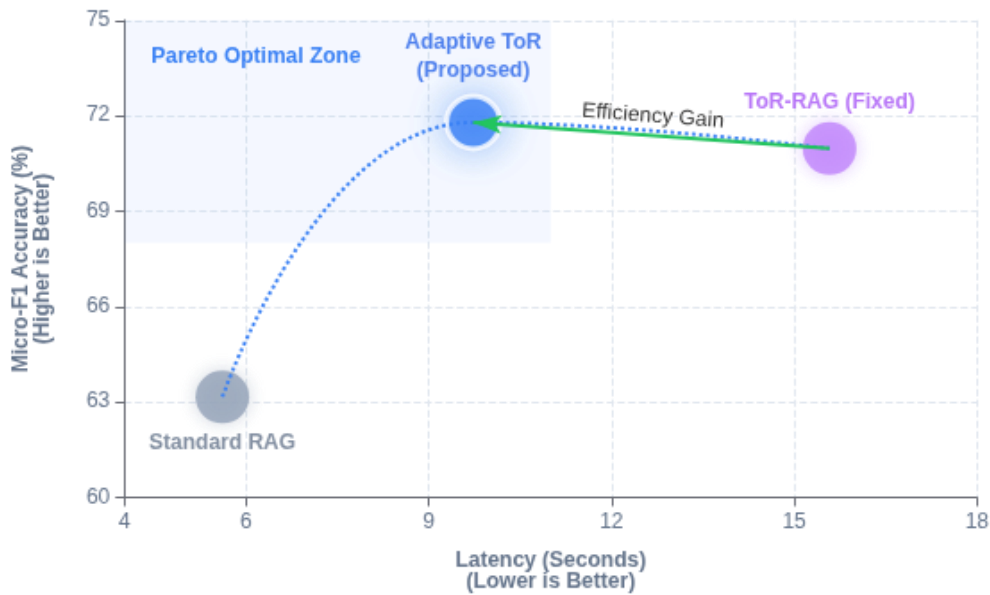

**Figure 5. Pareto Optimality Analysis**

Note. Pareto-optimality comparison of retrieval architectures on the NLU++ benchmark. The scatter plot visualizes the trade-off between Micro-F1 (y-axis) and query-processing latency (x-axis). Adaptive ToR (blue) occupies the Pareto-optimal frontier, achieving the highest Micro-F1 with moderate latency, thereby dominating both baselines. Fixed-depth ToR (purple) exhibits marginally lower Micro-F1 (70.97%) at substantially higher latency (15.58 s versus Adaptive ToR's 9.73 s), and Standard RAG (gray) achieves the fastest response (5.61 s) but at substantially lower Micro-F1 (63.16%). The blue shaded region represents the Pareto-optimal zone.

As reported in Tables 2 and 3, Adaptive ToR simultaneously achieves the highest accuracy scores and reduces latency, LLM invocations, and token consumption relative to Fixed-depth ToR-RAG. This combination of gains across all four dimensions establishes multi-objective Pareto dominance: no single dimension favors Fixed-depth ToR-RAG over Adaptive ToR.

The comparison with Standard RAG requires a more nuanced interpretation. Standard RAG achieves lower absolute latency (5.61 s versus 9.73 s), but this speed advantage comes at the cost of substantially lower accuracy ($-8.63$ pp Micro-F1, $-1.78$ pp Subset Accuracy). In application scenarios where multi-intent coverage is the primary requirement, this accuracy gap renders Standard RAG an inferior solution despite its latency advantage. Adaptive ToR thus occupies the Pareto-optimal frontier for workloads that prioritize intent classification quality.

The key enabler is complexity-aware adaptive depth control: by routing approximately 27% of queries through the fast single-step path (d=0) and reserving hierarchical decomposition only for genuinely complex cases (Section 5.2), Adaptive ToR eliminates the uniform over-processing that burdens fixed-depth architectures. This dynamic resource allocation is the mechanism by which the system achieves simultaneous improvements in accuracy, speed, and computational efficiency—a defining characteristic of Pareto-optimal design.

For reference, our concurrent work on ToR-Lite [8] demonstrates that a complementary Pareto position can be achieved on a different benchmark (MultiHop-RAG): ToR-Lite captures approximately 44% of Adaptive ToR's retrieval gain while operating at one-third the latency, delivering nearly twice the retrieval improvement per unit of computational cost—establishing that the Tree-of-Retrieval research lineage offers multiple Pareto-optimal configurations for different deployment constraints.

# 6. Conclusion

Adaptive ToR addresses a fundamental structural inefficiency in tree-based retrieval-augmented generation for multi-intent NLU: the indiscriminate application of uniform-depth hierarchical decomposition regardless of query complexity. By integrating a Query Tree Classifier (QTC) that computes a linguistically grounded Query Complexity Index (QCI), a Tree-Based Retrieval Module with adaptive depth control, an Adaptive Pruning Module (APM) employing two-stage precision-driven filtering, and a Retrieval Reranking Layer (RRL) with deduplication-first global rescoring, the proposed architecture dynamically configures retrieval topology a priori, enabling proactive resource allocation proportional to the intrinsic difficulty of each query.

Experimental results on the NLU++ benchmark validate that this complexity-aware dual-path design achieves Pareto-optimal performance across accuracy, latency, and computational efficiency dimensions simultaneously. Adaptive ToR attains 29.07% Subset Accuracy and 71.79% Micro-F1, representing a 9.7% relative improvement over fixed-depth ToR-RAG, while reducing query-processing latency by 37.6%, LLM invocations by 43.0%, and token consumption by 9.8%. The QTC effectively stratifies the query workload, routing 26.92% of queries through a fast single-step path (d=0) with 2.45 s response time, while token consumption scales 4.9-fold across depths, confirming that computational resources are allocated proportionally to query complexity.

These findings carry direct implications for adaptive information retrieval and intelligent NLU systems. The multi-objective Pareto dominance demonstrates that adaptive structural configuration, rather than post-hoc reranking, is the key mechanism for reconciling retrieval quality with computational efficiency. Sub-10-second average response times address established usability thresholds for interactive conversational systems, and depth-wise performance validation (Table 3) provides a quantitative mechanism study that informs the design of scalable, user-responsive NLU architectures.

The Tree-of-Retrieval research lineage now spans three complementary architectures: ToR-RAG [3] for fixed-depth hierarchical decomposition, Adaptive ToR for complexity-aware dynamic depth control, and ToR-Lite [8] for lightweight, LLM-free geometric decomposition. Together, these architectures offer multiple Pareto-optimal configurations along the accuracy-latency-cost frontier, enabling practitioners to select the appropriate architecture based on deployment constraints.

Future work will pursue four directions: (1) architectural lightweighting of the QTC for deployment in edge and on-device environments, (2) replacing the rule-based QCI with neural complexity estimators trained end-to-end on retrieval outcomes to improve cross-domain generalization, (3) extension to multimodal and cross-lingual retrieval settings to broaden the applicability of complexity-aware adaptive retrieval in diverse AI scenarios, and (4) integration of GraphRAG-based ontological knowledge structures into the tree retrieval framework. While Adaptive ToR currently decomposes queries into sub-queries without explicit domain knowledge, incorporating domain ontologies constructed via GraphRAG could enable semantically informed decomposition—guiding the tree expansion along ontologically grounded entity relationships rather than relying solely on LLM-generated sub-queries. This direction aims to reduce decomposition errors for domain-specialized queries (e.g., financial product hierarchies in Banking, service taxonomies in Hotel) while preserving the adaptive depth control that distinguishes Adaptive ToR from static graph-based approaches.